# Whale Optimization Based Energy-Efficient Cluster Head Selection Algorithm for Wireless Sensor Networks

Ashwin. R. Jadhav[1] and T. Shankar[2]

VIT University, Vellore, India
{[1]ashwinr.jadhav2013, [2]tshankar}@vit.ac.in

**Abstract.** Wireless Sensor Network (WSN) consists of many individual sensors that are deployed in the area of interest. These sensor nodes have major energy constraints as they are small and their battery can't be replaced. They collaborate together in order to gather, transmit and forward the sensed data to the base station. Consequently, data transmission is one of the biggest reasons for energy depletion in WSN. Clustering is one of the most effective techniques for energy efficient data transmission in WSN. In this paper, an energy efficient cluster head selection algorithm which is based on Whale Optimization Algorithm (WOA) called WOA-Clustering (WOA-C) is proposed. Accordingly, the proposed algorithm helps in selection of energy aware cluster heads based on a fitness function which considers the residual energy of the node and the sum of energy of adjacent nodes. The proposed algorithm is evaluated for network lifetime, energy efficiency, throughput and overall stability. Furthermore, the performance of WOA-C is evaluated against other standard contemporary routing protocols such as LEACH. Extensive simulations show the superior performance of the proposed algorithm in terms of residual energy, network lifetime and longer stability period.

## 1 Introduction

Wireless sensor networks (WSNs) is an emerging technology that has seen enduring ascent in its utilizations. This is due to its applications in surveillance, environment, and biosphere monitoring. Recent advances in the infrastructure systems have helped connect the world through the smart grid, intelligent networks, smart cities, and intelligent transportation [1]. These setups mentioned have a common component, the internet of things (IoT), where sensor nodes play a critical role, which is then closely coupled with information and communication technologies. An ordinary WSN consists of a large number of miniature, low-power, low-cost sensor nodes (SN) and at least one base station (BS) [2]. The objective of WSN is to screen at least one qualities of a specific range called the region of intrigue or the detecting zone. Correspondence range, adaptability, and energy efficiency are significant requirements of WSN. Sensors are for the most part furnished with non-rechargeable batteries, and once sent in the zone of intrigue they continue working until they come up short on power. The sensor nodes are frequently deployed in the remote or hostile environment, so replacement of their battery is not feasible. WSNs use their energy in several ways, yet data transmission is one of the biggest causes for energy expenditure. Consequently, the design of an energy-efficient protocol for WSN is one of the biggest difficulties in increasing the system's life expectancy. Besides, the lifetime can be influenced by system scale. As the scale builds, the dependability of the system turns out to be critical. Diverse strategies have been recommended for prolonging the lifetime of the sensor network. Clustering is one of the most proficient ways to deal with energy utilization in WSN [3], and a hierarchical architecture design gives a strong answer for the network lifetime and stability issues. In a hierarchical architecture design, the network is divided into various layers and different nodes perform distinctive tasks. The sensing area of the WSN is partitioned into several groups, called clusters. Each of these clusters consists of cluster members (CM) and a single cluster head (CH) (see Fig. 1). First, the cluster head (CH) gathers the data from every cluster member and aggregates the gathered data. Next, it sends the aggregated data to the remote base station (BS). In addition, the BS is connected to a private network or a public network. This approach of clustering has numerous benefits. First, it helps in discarding redundant data at the CH level itself. Thereby decreasing the energy consumption of each node due to the reduced data transmission. Second, the network's scalability is increased significantly. Finally, it also conserves the bandwidth as the sensor nodes communicate directly to their cluster head and this decreases the number of redundant messages amongst the sensor nodes

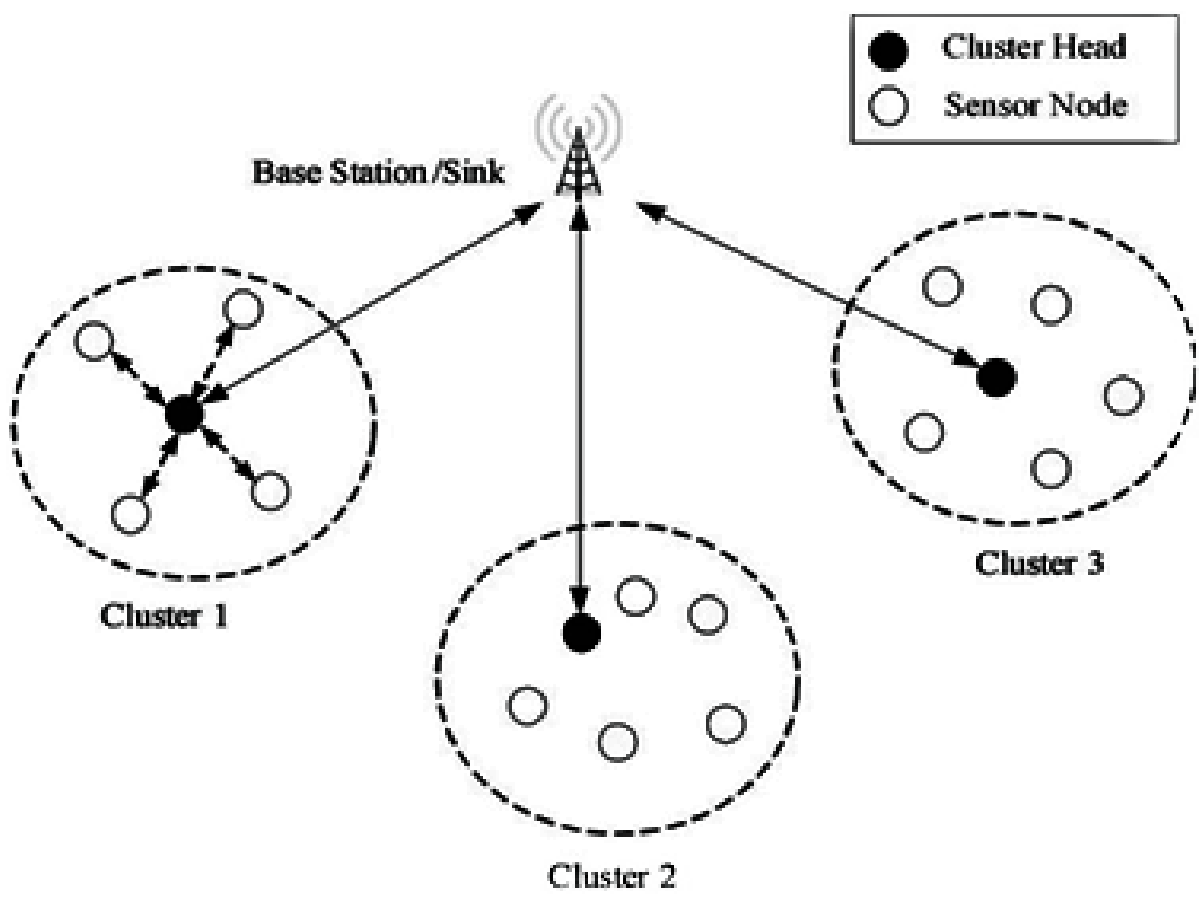

**Fig. 1.** Clustered hierarchical architecture

CH-selection is an optimization problem and turns out to be a NP-hard problem as well [37]. Canonical optimization techniques are ineffective when the network is scaled up. Whale optimization algorithm (WOA) is bio-inspired optimization algorithm which is effective for NP-hard problems. Furthermore, it is easy to implement, gives high-quality solutions, and has commendable exploration and exploitation features which lead to quick convergence with the ability to escape from local minima.

In this paper, the main focus is on the CH selection issue and a WOA based algorithm is presented called Whale optimization algorithm based clustering (WOA-C). The algorithm helps select the CHs among the sensor nodes efficiently taking into account the residual energy of the nodes. Algorithm is evaluated extensively to show its superiority over contemporary routing techniques.

The rest of the paper is organized as follows. Section 2 follows up on related works. Section 3 presents Whale optimization algorithm (WOA). A detailed explanation of the proposed algorithm is given in Sect. 4. The performance evaluation methods are provided in Sect. 5. Finally, the simulation results are explained in Sect. 6 followed by the conclusion in Sect. 7.

## 2 Review of Related Literature

Clustering approach is very important in WSN, at the same time it can be very complex. Consequently, numerous clustering algorithms have been developed for WSNs. There have been different kinds of approaches to the problem of clustering with the main two techniques discussed being heuristic and nature-inspired approaches. However, the proposed algorithm is based on nature-inspired approach and hence, more emphasis is given to that section.

### 2.1 Heuristic Approaches

A substantial number of clustering algorithms [4–7] based on heuristic strategies have been created for WSNs. Amongst this LEACH [4] is a popular clustering algorithm in which the sensor nodes elect themselves as a CH with some likelihood. LEACH provides compelling energy savings and thus, prolongs the lifespan of the system contrasted with static clustering algorithms and minimum transmission energy (MTE) protocol. On the other hand, the fundamental disadvantage of this algorithm is that the CH selection is random. Consequently, there is a plausibility to choose a CH with low energy, which can die quickly and end up reducing the performance of the network. Therefore, a variety of algorithms were developed to enhance LEACH protocol.

Consequently, PEGASIS [8] is the other prominent protocol that was later developed. PEGASIS sorts out the nodes into the chain so that every node transmits and gets the information solely from its neighbor hubs. In each cycle, an arbitrarily chosen node from the chain is designated as a CH. PEGASIS is more efficient in terms of energy consumption although it quickly becomes unstable as the network is scaled up. As of late, numerous algorithms [9–13] have been produced for data congregation schemes for augmenting the lifetime of WSNs. A gateway based energy-aware multi-hop routing protocol (M-GEAR) is proposed by the authors of [14]. The coverage area is coherently partitioned into four sectors. Furthermore, the nodes in the primary range convey specifically to the base station while nodes in different sectors corre-spond to the gateway node. Nodes in other two sectors use clustering hierarchy and sensor nodes transmit their information to gateway portal from their CHs.

M-LEACH is an algorithm proposed by Xiaoyan et al. [15], it builds upon LEACH with the main distinction being, as opposed to sending the information straightforwardly to the BS, it forwards to the following subsequent hop CH node. The V-LEACH protocol was proposed by Yassein et al. [16] for enhancing the LEACH protocol. The main idea is to select some vice-CHs along with main CHs so that when primary CH dies, the vice-CHs can take over as the main CH. It is shown to have superior performance over LEACH but at the same time, the sensor nodes require additional processing energy for choosing vice CHs. The E-LEACH protocol was proposed by the authors of [17] to further improve LEACH protocol. It involves selection of CHs while taking into account of the residual energy of each node which can extend the lifetime of the network but avoid the selection of a node with low residual energy. As indicated by [18], M-LEACH is superior to TL-LEACH while E-LEACH [17] is superior to M-LEACH. The authors of [19–22] have proposed energy proficient cluster-based routing protocols for the stable network. Furthermore, [23] provides techniques for analysis of energy in WSNs. The authors of [24, 25] propose algorithms for topology management and delay tolerance networks. In order to address the issues of network security [26] have proposed algorithms to make WSN more secure.

### 2.2 Nature-Inspired Approaches

Numerous clustering algorithms were developed based on utilization of nature inspired algorithms in light of advanced and more efficient nature-inspired approaches being developed. The LEACH-C was proposed in [27] to further develop LEACH protocol. In LEACH-C, the cluster formation is done toward the start of each round utilizing a centralized algorithmic program by the base station. The base station utilizes the data gotten from every sensor amid the setup stage to locate a foreordained number of cluster heads such that the clusters formed help minimize the energy consumption of non-cluster nodes. The downside of this protocol is that it completely overlooks the arrangement of clusters, which leads to energy inefficiency of the network.

In order to eliminate the limitations of previously mentioned algorithms, metaheuristic algorithms have been favored for the ideal selection of efficient cluster heads. Every meta-heuristics algorithm has two main segments, to be specific, diversification and intensification. In addition, in order to be efficient, the algorithm must have the capacity to obtain a diversified range of solutions with includes the best optimal solution in order to investigate the entire search space while intensifying the search in the neighborhood of optimal solution. Keeping in mind the end goal to do as such, the entire hyperspace must be accessible, though, not every part is visited throughout the search.

As the solution to energy efficiency in WSNs becomes more complex and computationally expensive with every newly developed routing protocols, more emphasis has been given to bio-inspired metaheuristic algorithms such as Artificial Bee Colony

Optimization (ABC) [28], [29], Ant Colony Optimization (ACO) [30-32], Fuzzy Logic [33], [34] and Particle Swarm Optimization (PSO) [35], [36]. An energy aware cluster head optimization was developed by Latiff et al. [37] using PSO called PSO Clustering (PSO- C). While it considers meaning inter-cluster distances and ratio of the initial energy of all nodes to the current residual energy of all nodes, it fails to consider sink distance which is important for communication between CHs and BS. Hybrid metaheuristics algorithms have also been implemented. The authors [38] developed hybrid PSO-HSA for efficient cluster head selection. The Whale Optimization Algorithm (WOA) algorithm [39] is a new evolutionary optimizing algorithm that can be used for optimization problems; it has demonstrated to have on par performance with PSO and genetic algorithms.

In this paper, a centralized, energy aware cluster-based algorithm called WOA clustering (WOA-C) is developed in order to extend the lifetime of a sensor network by using WOA algorithm. The algorithm is designed in such a way so as to consider the use of a high-energy node as a cluster head and form clusters that are uniformly distributed throughout the entire sensor plot. Ultimately, the primary idea of the proposed algorithm is the selection of an optimal cluster head that limits the intra-cluster distance between itself and member nodes, and the enhancement of energy utilization of the network.

## 3 Whale Optimization Algorithm

Whale Optimization Algorithm (WOA) is a nature inspired meta-heuristic optimization algorithm introduced by Mirjalili et al.[39]. WOA is based on the hunting behavior of humpback whales. Furthermore, WOA simulates the hunting behavior with random or the optimal search agent to hunt the prey (exploration) and the use of a spiral bubble-net attacking mechanism of humpback whales to simulate the catching of prey (exploitation).

Nature-inspired metaheuristic algorithms have been effective for optimization problems and help in find the optimal solution. The search process of most metaheuristic shares a common feature. It involves two phases: exploitation and exploration. Whale Optimization Algorithm (WOA) is inspired by the survival and hunting behavior of humpback whales. Whales can survive alone or in groups and can be up to 30m in length. Moreover, humpback whales have a unique hunting method called bubble-net feeding method which usually involves creating bubbles along a circle around the prey while hovering around the prey. Usually, there are two maneuvers associated with this hunting technique. First one is ‘upward-spirals’, where the whale dives 12m down and creates bubbles in spiral shape while swimming towards the surface and the other one is more complex and has three stages namely, lobtail, capture loop, and coral loop. This unique spiral bubble-net hunting behavior can only be seen in humpback whales.

Just like every optimization technique, WOA involves two phases: exploitation and exploration. Exploration refers to a global search for optimal solutions, whereas exploitation is related to local search. Exploitation consists of probing a constrained (yet promising) region of the search space with the hope of enhancing a good solution ‘S’ that is already known. This operation amounts then to intensifying (refining) the search in the vicinity of that solution ‘S’. In other words, local search is being conducted. Meanwhile exploration, on the other hand, consists of probing a much greater region of the search space with the expectation of discovering other encouraging solutions that are yet to be refined. This operation amounts then to diversifying the search so as to avoid getting trapped in a local optimum. Ultimately it is similar to global search. In WOA the hunting is analogous to optimization technique and the location of prey is analogous to the location of the best solution. In addition, the WOA algorithm starts with a randomly generated population of whales (solutions) each with the random position. In the first iteration, the search agents update their positions in reference to a randomly chosen search agent. However, from second iteration onwards the search agents update their position with respect to the best solution obtained so far. A random search agent is chosen if the value of |A| > 1, this helps in exploration. When the best solution is selected, |A| is set to |A| < 1. This induces exploitation as all the search agents will converge. Therefore, WOA can be considered as a good global optimizer. The hunting behavior can be explained in 3 phases: searching, encircling, and attacking the prey.

### 3.1 Encircling of Prey

Whales update their positions imitating the encircling behavior during the optimization, around the position of the current best search agent. The encircling is mathematically modeled by the following equations:

$$\vec{D} = |\vec{C} \cdot \vec{X}^{*}(t) - \vec{X}(t)| \quad (1)$$

$$\vec{X}(t+1) = \vec{X}^{*}(t) - \vec{A} \cdot \vec{D} \quad (2)$$

Where t is the current iteration, $\vec{A}$ and $\vec{C}$ are coefficient vectors and $\vec{X}$ is the position vector. The position vector of the best solution that has been obtained so far is given by $\vec{X}^{*}$. The value of $\vec{X}*$ is updated after each iteration if a better solution exists. The coefficient vectors $\vec{A}$ and $\vec{C}$ are calculated by Eq. 3 and Eq. 4:

$$\vec{A} = 2\vec{a} \cdot \vec{r} - \vec{a} \quad (3)$$

$$\vec{C} = 2 \cdot \vec{r} \quad (4)$$

Where vector $\vec{a}$ linearly decreases from 2 to 0 over the course of iterations and $\vec{r}$ is a random vector that varies between [0, 1].

### 3.2 Attacking of Prey (Exploitation Phase)

The attacking behavior is modeled with respect to the bubble net attacking strategy. Two main approaches are taken in order to model the bubble-net behavior of humpback whales. The two methods are shrinking encircling mechanism and spiral updating position mechanism as given by Fig. 2 [39]. Humpback whales can execute any of this two mechanism to catch the prey and therefore, each of this mechanism can happen with a probability of 50%. A random variable p is introduced where p varies between [0, 1].

In shrinking encircling mechanism the value of $\vec{a}$ is decreased in Eq. 3. The value of A is a random value between [-a, a] where $\vec{a}$ is decreased from two to zero over the total number of iterations. The spiral updating position mechanism is done in such a way so as to mimic the helix structured maneuver of humpback whale. The attacking of prey is the exploitation phase of this optimization algorithm. . Exploitation consists of searching a restricted (yet promising) region of the search space with the hope of enhancing the within the neighborhood of solution ‘S’. This operation amounts then to intensifying (refining) the search in the vicinity of that solution ‘S’. The value of $\vec{A}$ is between [-1, 1] and therefore when $|\vec{A}| < 1$, then exploitation is induced and all the search agents converge to obtain the best solution. The updating model can be given by Eq. 5.

$$\vec{X}(t+1) = \begin{cases} \vec{X^*}(t) - \vec{A} \cdot \vec{D} & if \;\; p < 0.5 \\ \vec{D'} \cdot e^{bl} \cdot \cos(2\pi l) + \vec{X^*}(t) & if \;\; p \geq 0.5 \end{cases} \tag{5}$$

### 3.3 Searching of Prey (Exploration Phase)

The exploration phase is based on the variation of vector A and it is to mobilize the search agents in search of better solutions similar to a global search. The value of |A| is set to greater than 1 which forces the search agent to diverge far away in the search space. In stark contrast to the exploitation phase, the search agents update their position with respect to randomly chosen search agent instead of referring to the best search agent. The searching mechanism can be modeled by Eq. 6 and Eq. 7.

$$\vec{D} = |\vec{C} \cdot \vec{X}_{rand} - \vec{X}| \tag{6}$$

$$\vec{X}(t+1) = \vec{X}_{rand} - \vec{A} \cdot \vec{D} \tag{7}$$

In this paper, the WOA algorithm is used for the cluster head selection process. These equations and update mechanisms will later be used in the cluster head section.

# 4 Proposed Algorithm

A centralized, energy aware cluster-based algorithm called WOA-Clustering (WOA-C) is developed in this work. The algorithm is designed in such a way so as to consider the use of a high-energy node as a cluster head and form clusters that are uniformly distributed throughout the entire sensor plot. The proposed algorithm has three main components, it includes network model, energy model and cluster head selection model which are explained in the following sections.

The network model has some important assumptions about the sensor nodes, plot area, and other constraints. The energy model is important for establishing the value of energy dissipation for sensor nodes while transmitting or receiving data. While the cluster head selection gives details about how WOA is used for our problem.

## 4.1 Network Model

The network model is considered to be free space model. It has a transmitter and a receiver with a distance of separation, d. The amplifier circuits are also present at both Tx and Rx. The following properties about the WSN are assumed:

- All sensor nodes are randomly deployed and are stationary.
- All nodes are homogeneous and have limited energy.
- The BS is stationary and it can be located inside or outside the sensing area.
- Every node gathers the data periodically and always has some data to forward.
- Nodes don't know their exact positions nor the position of other nodes.
- The nodes are self-organizing and need not be monitored after deployment.
- Data fusion is used to minimize the total amount of forwarded data.
- Every node has the capability to operate as a cluster head.

The WSN scenario considered for simulation has all the above properties and limitations. The nodes can compute the distance between the BS and other nodes by comparing the received signal strength. Hence, it doesn't need any additional system with location services such as GPS. Also, a node joins the cluster whose CH is nearest to it.

## 4.2 Energy Model

The energy model in this paper is based on the first order radio model as used in [4]. In this particular model, the transmitter needs to drive radio electronics and power amplifier which consumes energy. Likewise, the receiver dissipates energy to drive the radio electronics on its side. Furthermore, the energy consumption of the node is proportional to the quantity of the data and distance over which it has to be sent. Since the attenuation with distance, an energy loss model with $d^2_{ij}$ is used for relatively short distances while $d^4_{ij}$ is used for longer distances, where $d_{ij}$ is the distance between sensor nodes *i* and *j*. For this reason, the propagation distance (d) is compared to a threshold distance $d_0$ and when propagation distance (d) is lesser than $d_0$ energy con-sumption of a node is proportional to $d^2$ otherwise it is proportional to $d^4$ [4]. The total energy consumed by a node for transmitting $l$-bit data packet is given by Eq. 8.

$$E_{TX}(l, d) = l \cdot E_{elec} + l \cdot E_{elec}\, \varepsilon_{fs}\, d^2, \qquad \text{if } d < d_0$$
$$= l \cdot E_{elec} + l \cdot E_{elec}\, \varepsilon_{mp}\, d^4, \qquad \text{if } d > d_0 \tag{8}$$

Where $E_{elec}$ is the energy dissipation per bit for transmitter or receiver circuit. $E_{elec}$ relies on numerous elements such as the digital coding, modulation technique, and spreading of the signal. Amplifier energy for free space is $\varepsilon_{fs}$ and for multi-path model, it is $\varepsilon_{mp}$ which depends on transmitter amplifier model. The energy consumed by the receiver while receiving l-bit of data is given by Eq. 9.

$$E_{RX}(l) = l \cdot E_{elec} \tag{9}$$

### 4.3 Cluster Head Selection

The operation of the proposed algorithm is based on a centrally controlled algorithm that is enforced from the base station (BS). The proposed algorithm operates in rounds, in which every round starts with a setup phase similar to the approach in [2 PSO-C]. At the beginning of every setup phase, all nodes transmit information regarding their current energy status and location to the base station. With this information, the base station computes the mean residual energy and then ensures that only the nodes with more residual energy than the mean energy value will be eligible to be selected as the cluster head. Then, the base station then runs the WOA algorithm to determine the 10 best cluster heads that minimize the fitness function.

The proposed WOA implementation is on the randomly deployed stationary nodes in the sensor network. It assumed that n nodes represent the cluster head search agents (whales), which is ($CH = CH_1, CH_2, \ldots, CH_n$). For imitating the positions of the search agents (whales) in WOA, and since sensor nodes are stationary, the position of the search agent (candidate CH) is represented by $CH_i$ in the 2D space indicating the node's positions [$Pos_i(t) = x_i(t), y_i(t)$]. The best search agent's position is then used to determine the best solution, which is used to select the optimal cluster heads. The pseudo code of the proposed algorithm is given below:

The search agent is initially positioned randomly and then clones the nearest node to its position. The fitness value is calculated for all the search agent and the best one is selected for reference. The parameters of WOA is updated such that the other search agents position themselves with respect to the best agent. The flowchart of the entire process is given by Fig. 3.

The cluster head (CH) choice is dictated by a fitness function. In the WOA optimization, the fitness function plays a critical role in the exploration of prey component. The contribution for this is the node's attributes, which includes its remaining residual energy ($Er$) and number of neighbors.

The fitness function is given by Eq. 10 [40]:

$$f\,(CH_i) = p_1\,|\,N\,(CH_i)\,| + p_2\,\Sigma\,(CH_E) \tag{10}$$

Where the p1 and p2 are randomly chosen between 0 and 1. $N\ (CH_i)$ is the list of all the node neighbors around a particular cluster head $CH_i$, and the $CH_E$ is the neighbor node's residual energy level. The best solution is the one with the largest fitness value and hence it will have enough residual energy and adequate number of neighboring nodes in order to become the CH.

**Algorithm 1** WOA-BASED CH SELECTION

```
while r < rmax do                                        [ For each round r ]
  while t < tmax do
      for each Whale (search agent) CHi do:
            Clone the nearest node to CHi
            Compute fitness according to Eq. 10
            Calculate the coefficient vectors according to Eq.: (1-7)
            Update the positions of the Whales
            Update the X* best position if X is better than X*
            Update the fitness function for all search agents
      end for
  end while
CH = Nearest node to the X* position
end while
```

**Fig. 2.** Shrinking encircling mechanism (a) and Spiral updating position (b)

After the base station has distinguished the optimal arrangement of cluster heads and their related group nodes, the base station transmits the data that contains the CH-ID toward every node in the network. The selected cluster head becomes a control center in its neighborhood in order to organize the transmission of data. Moreover, a TDMA (Time Division Multiple Access) schedule is arranged for its members in order to avoid collision of data amongst the cluster members. Furthermore, this enables the cluster members to be in sleep-wake cycle where, they need to be awake only in their respective TDMA time slots and conserve energy during its sleep cycle. In sleep cycle, the nodes can be in low power state and conserve energy. After receiving data from all the cluster members at the end of each round, the cluster head then combines all the data and transmits that fused data back to the base station. The data between BS and CH is done using CDMA (Carrier Sense Multiple Access) approaches which is similar to [2].

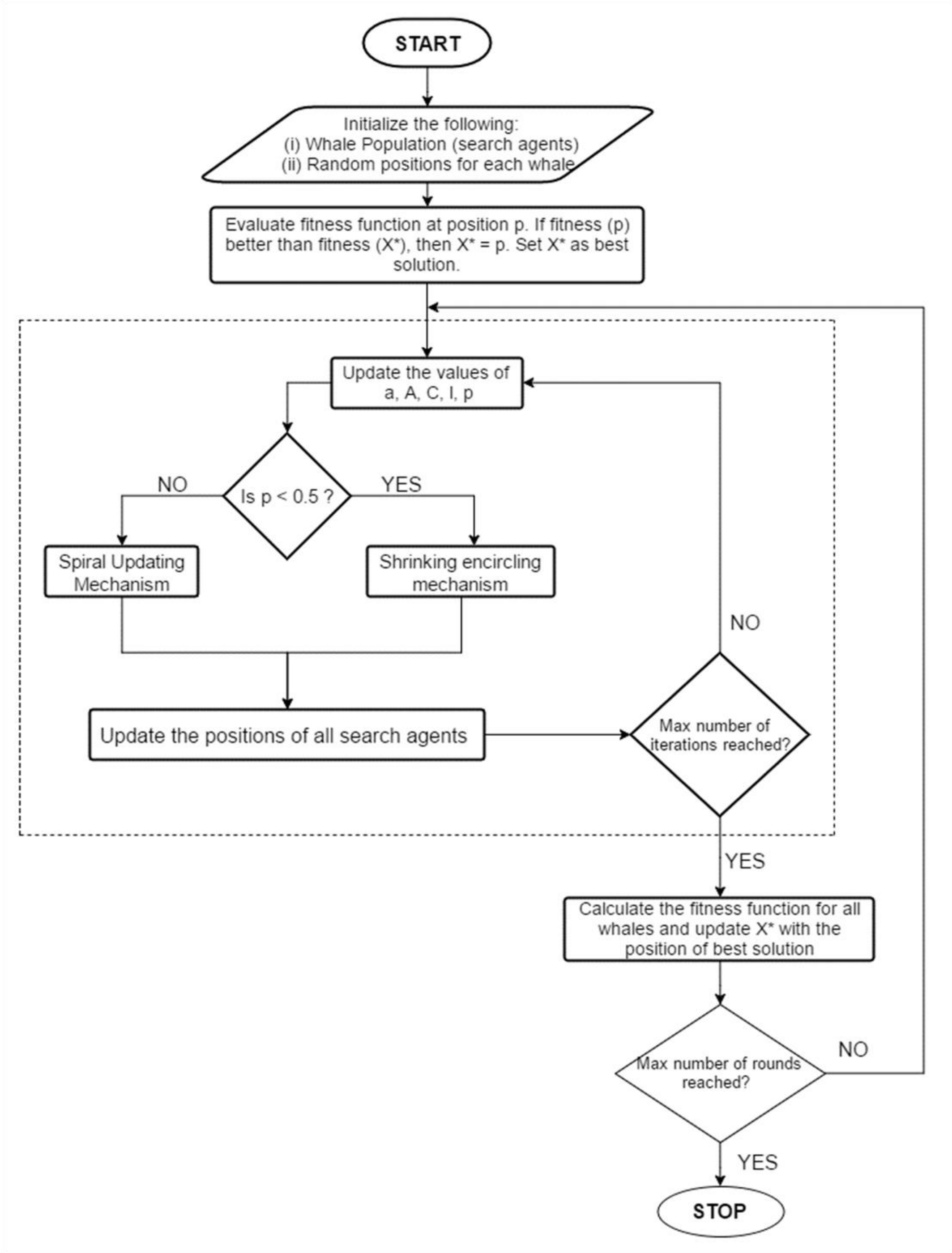

**Fig. 3.** Flowchart of WOA algorithm for CH selection

# 5 Performance evaluation

## 5.1 Simulation environment

The proposed algorithm was simulated in MATLAB 2016b and results were plotted. The system configuration was Intel Core i7-4710 HQ processor, 8 GB RAM running Windows 10 and with 4 GB Nvidia 860M GPU. The simulations were performed many times with varying conditions such as a number of nodes and number of cluster heads. The network sensing area was assumed to be 100 x 100 $m^2$. The initial simulations are on WSN #1 with 100 nodes and 10 cluster heads. Then, the simulations were also performed on WSN #2 with 300 nodes, 30 CHs and the third scenario WSN #3 with 500 nodes and 50 CHs. The base station (BS) position was also varied for different scenarios. In the first scenario the base station was positioned inside the sensing area at (50x50), then in the second scenario the base station is positioned at the edge of sensing area at (100x100) and finally, in the third scenario the base station is positioned outside the sensing area at (50x200). The algorithm was run for 20 times and the average of the instances of the resultant data was chosen for plotting results. The algorithm was tested with a predefined search agent's population of 30. The various parameters considered for simulations are given in Table 1.

**Table 1.** Network Parameters

| **Parameter** | **Value** |
|---|---|
| Target area | 100 x 100 $m^2$ |
| Base Station Position | (50-100, 50-200) |
| Number of Nodes | 100-500 |
| Initial Energy of node | 0.5 J |
| Transmitter/Receiver Electronics - $E_{elec}$ | 50 $nJ/bit$ |
| Transmit amplifier (free space) - $\epsilon_{fs}$ | 10 $pj/bit/m^2$ |
| Transmit amplifier (multipath) - $\epsilon_{mp}$ | 0.0013 $pj/bit/m^4$ |
| Data Aggregation Energy cost | 5 nJ/ bit |
| Packet Size | 4000 bits |
| Message Size | 200 bits |
| $d_0$ | 30m |
| Number of agents | 30 |
| Number of Iterations | 500 |
| Fitness function probability - p1 | 0.7 |
| Fitness function probability - p2 | 0.3 |

### 5.2 Performance evaluation metrics

In order to measure the performance of the proposed algorithm, the following metrics have been used.

1. *Energy Utilization* - The total energy consumption at a given round can give a good estimate of the energy efficiency of the algorithm and the total energy consumption increases with the increasing number of rounds. Note that this also applies to residual energy of the network and the same estimates can be drawn based on that
2. *Network Lifetime* - The network lifetime is defined as the total number of rounds until which the last node is alive. Last node death (LND) can be noted by plotting number of dead nodes vs the number of rounds. Network lifetime increases with increase in energy efficiency.
3. *Network Throughput* - The throughput of the network gives an estimate of the amount of useful data being received by the BS. Throughput of the network is noted every round and is plotted. Hence, network throughput is an important criterion for any routing algorithm.

### 5.3 Simulations and Analysis

The simulations were done based on many parameters and they were done under varying conditions. The results and analysis of these simulations are given in the subsequent sections.

**Energy consumption performance:** The algorithm was run under different conditions. The total number of nodes varied between 100-500 and the number of CH varied from 10 to 50. There were three different scenarios, WSN#1 with 100 nodes and 10 cluster heads, WSN#2 with 300 nodes and 30 cluster heads, and finally WSN#3 with 500 nodes and 50 cluster heads. Other routing protocols were also tested under similar conditions for performance comparisons. Direct transmission (DT), LEACH, LEACH-C, PSO-C were compared with the proposed algorithm WOA-C with WSN#1 as the default wireless sensor network scenario. Here, the energy consumption of different routing methods is compared. Fig. 4 gives the residual energy of the sensor network using different routing techniques in WSN#1 with 100 nodes and 10 CHs, with a central BS position at (50x50).

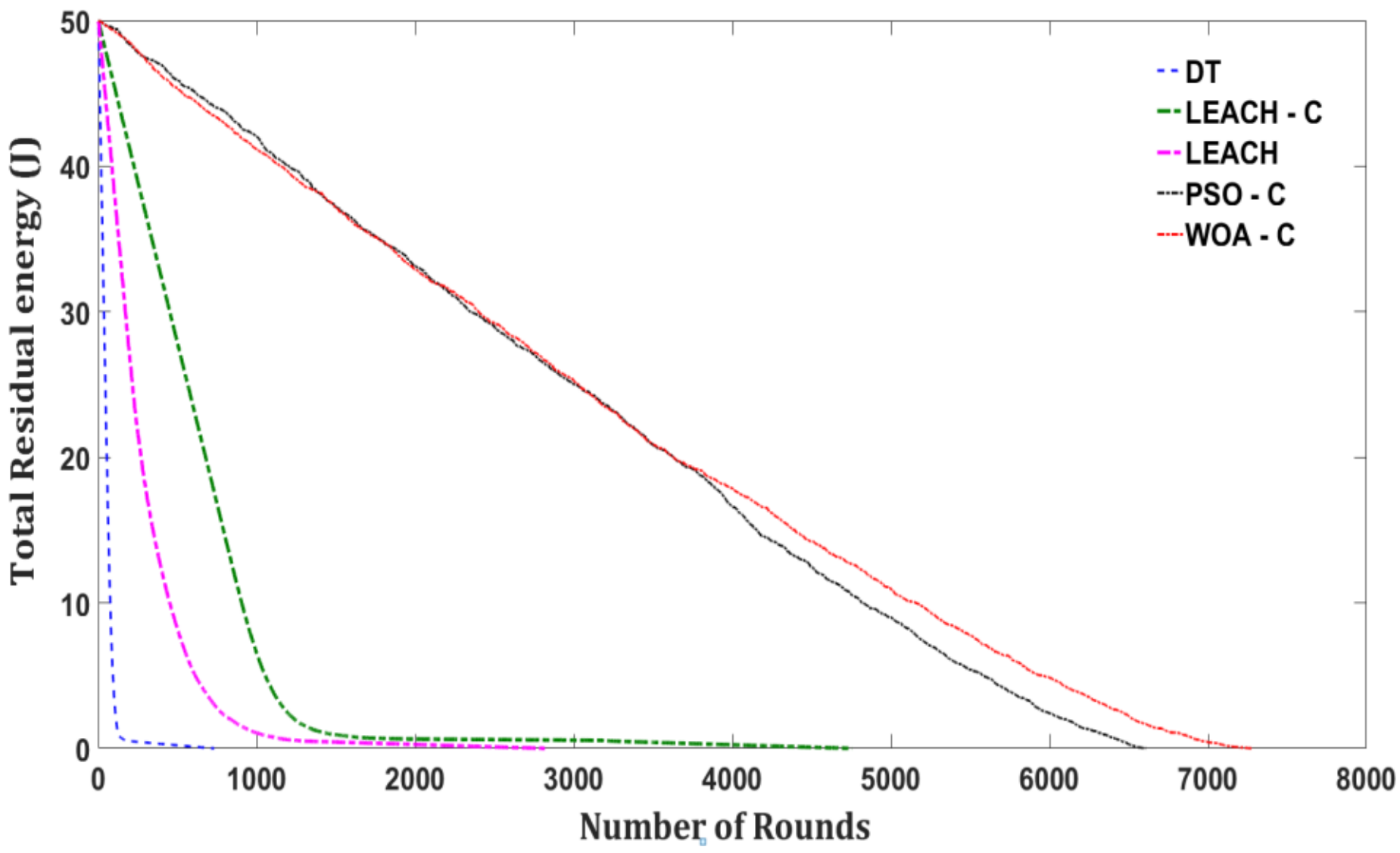

**Fig. 4.** Comparison in terms of total residual energy in WSN#1 with 100 nodes and 10CHs

The proposed algorithm performs better because of the fact that the proposed algorithm considers the energy of the node before selecting it as CH. Also, the nodes transmit to the nearest CH and consume less energy as a result.

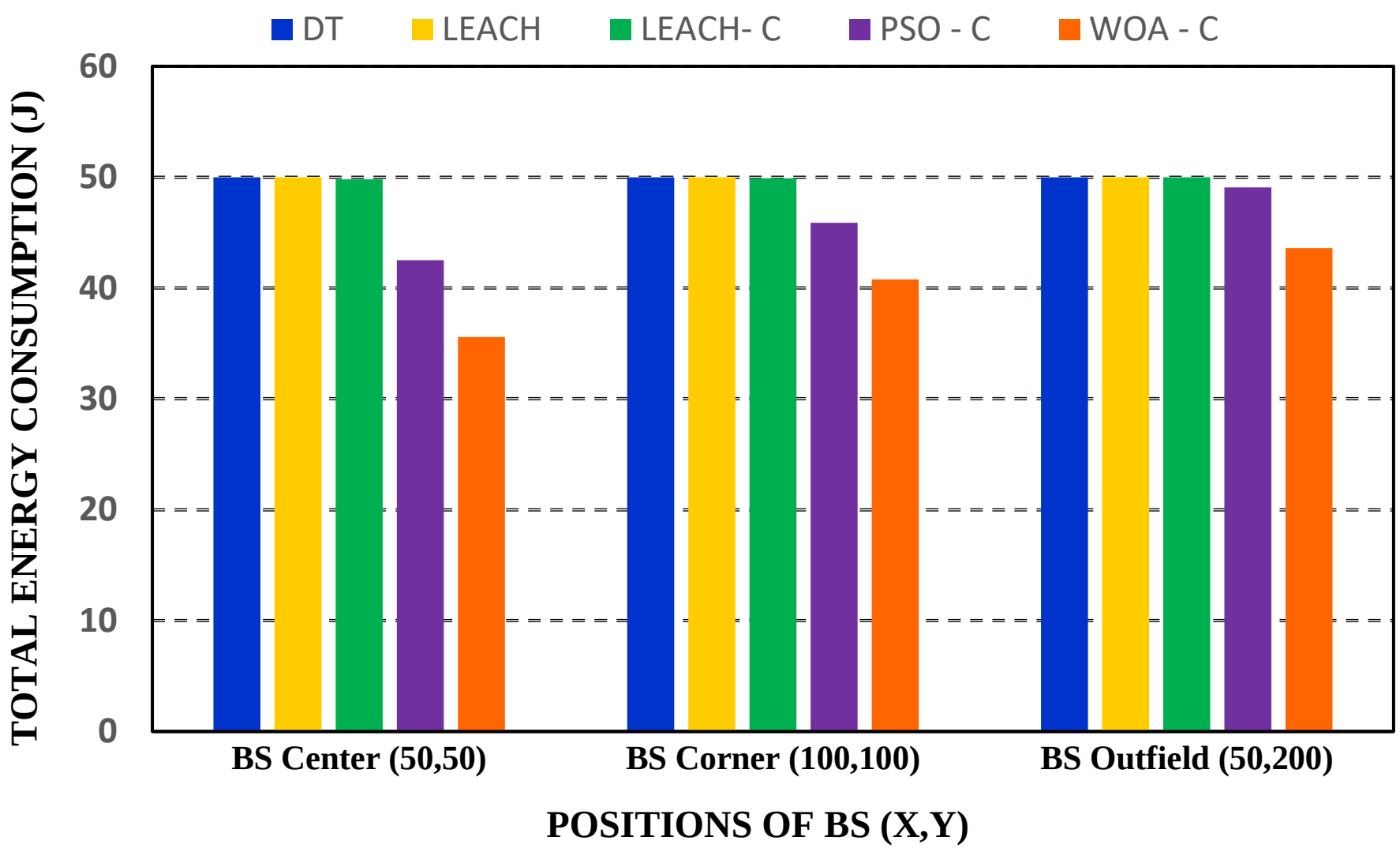

**Fig. 5.** Comparisons of energy consumption with varying BS positions in WSN#1 with 10 CHs.

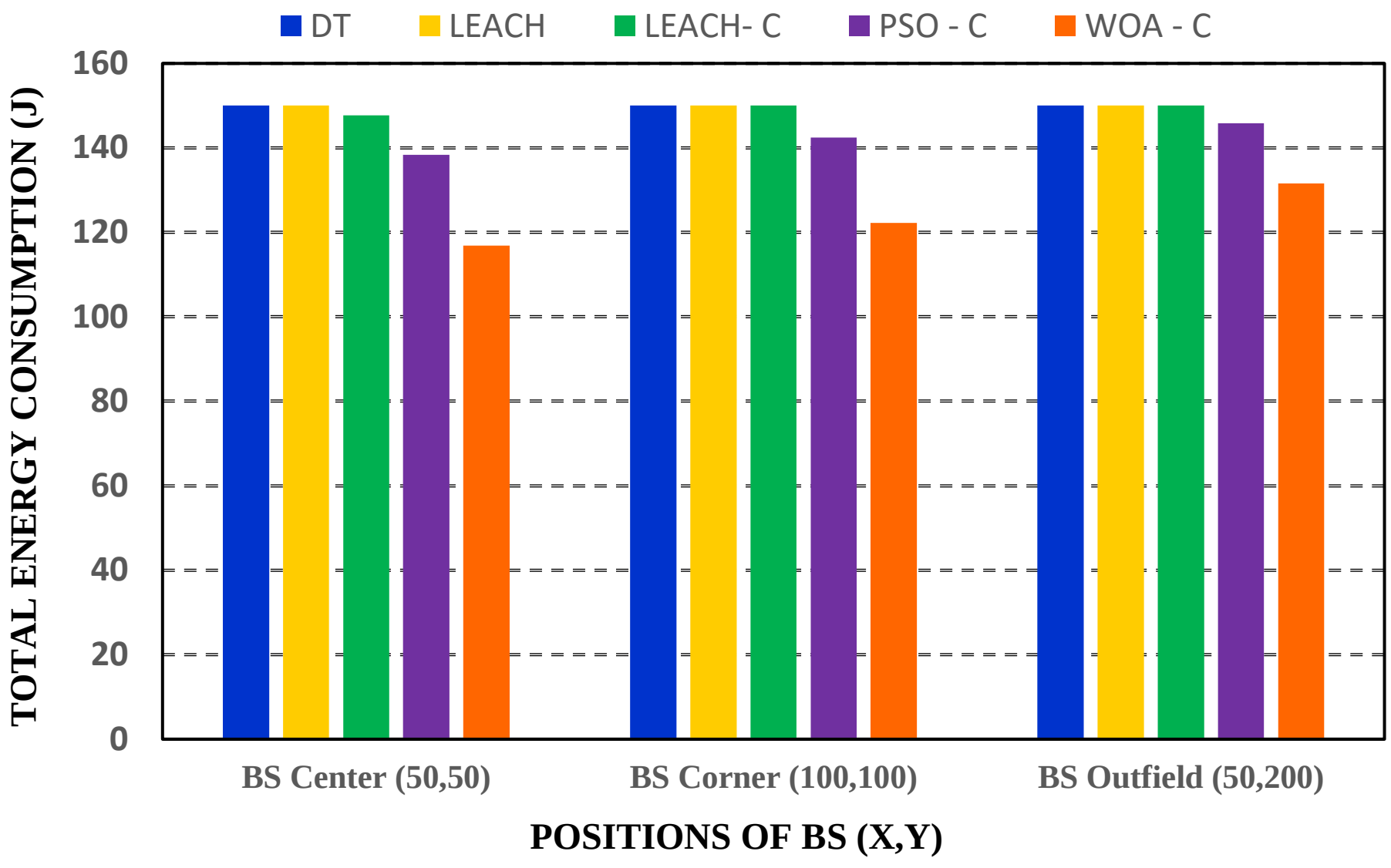

**Fig. 6.** Comparisons of energy consumption with varying BS positions in WSN#2 with 30 CHs.

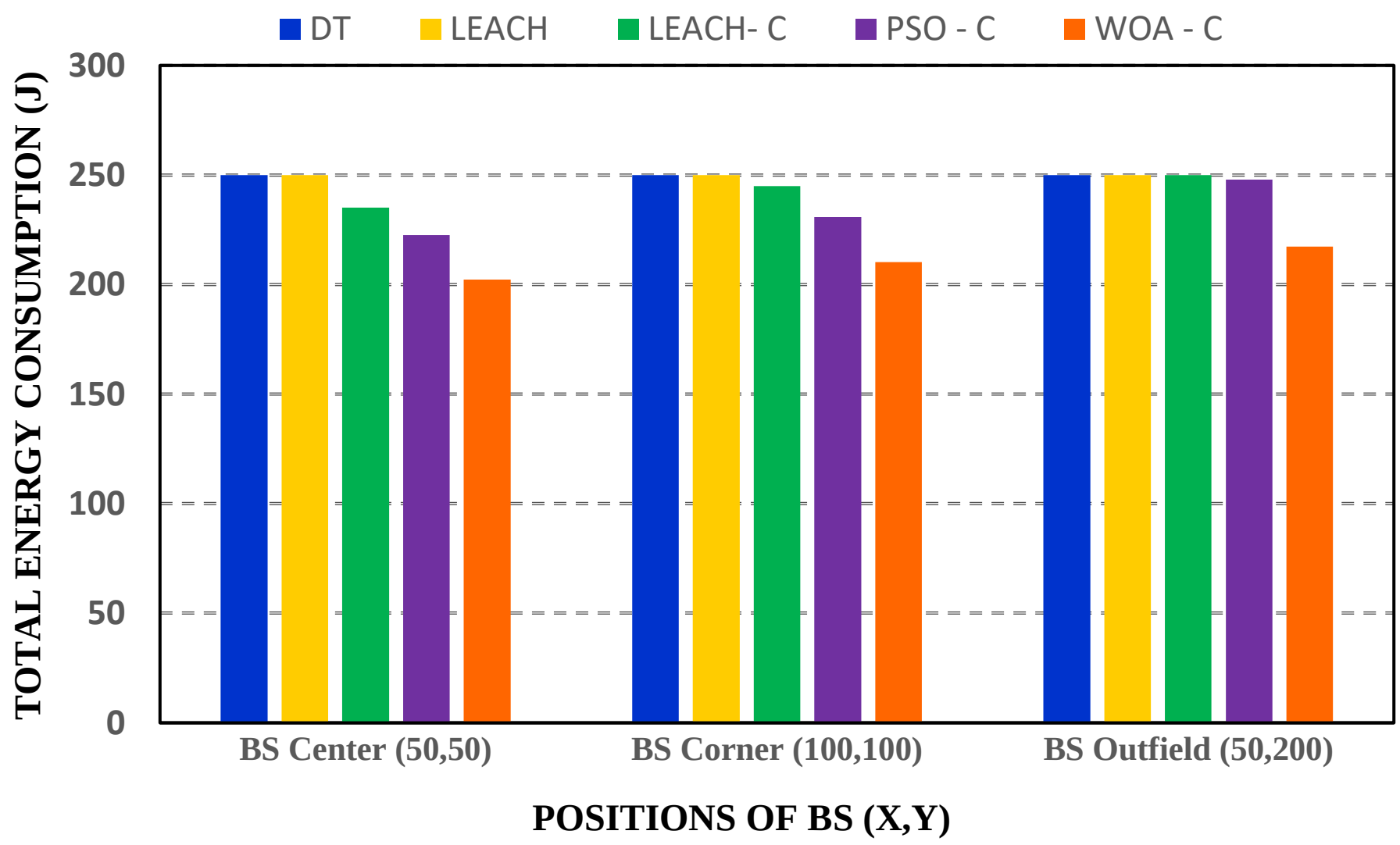

**Fig. 7.** Comparisons of energy consumption with varying BS positions in WSN#3 with 50 CHs.

Furthermore, the simulations are done for bigger networks with more cluster heads. The base station (BS) position is also varied from a central location of (50x50) to corner position of (100x100) and outfield location of (50x200). The algorithms were run to compare energy consumption across different scenarios. The sensor nodes were varied from 100 to 500 with the cluster heads varied from 10 to 50 accordingly. Fig.5, 6, 7 show the total energy consumed by various algorithms under varying conditions at round 5000.

The energy consumption in each scenario varies for different algorithms. As the size of the network increases, the energy performance of LEACH, LEACH-C and PSO-C diminishes. The proposed algorithm however, performs much better as the number of nodes increase. The proposed algorithm outperforms all other routing protocols when it comes to stability and independence from BS position. Often in real time applications, the BS position is not known beforehand; hence, an algorithm whose performance is independent of BS position is highly preferred which is exactly what the proposed algorithm offers. The energy consumption values are also given in Tables 2, 3, 4.

**Table 2.** Total energy consumed at round 5000 in WSN #1 with 10 CHs.

| Number of Sensors = 100 | BS center | BS corner | BS outfield |
|---|---|---|---|
| Direct Transmission (DT) | 50.00 | 50.00 | 50.00 |
| LEACH | 50.00 | 50.00 | 50.00 |
| LEACH -C | 50.00 | 50.00 | 50.00 |
| PSO - C | 42.53 | 45.86 | 49.07 |
| WOA - C | 36.58 | 40.76 | 43.58 |

**Table 3.** Total energy consumed at round 5000 in WSN #2 with 30 CHs.

| Number of Sensors = 300 | BS center | BS corner | BS outfield |
|---|---|---|---|
| Direct Transmission (DT) | 150.00 | 150.00 | 150.00 |
| LEACH | 150.00 | 150.00 | 150.00 |
| LEACH -C | 147.76 | 150.00 | 150.00 |
| PSO - C | 138.36 | 142.56 | 145.87 |
| WOA - C | 118.58 | 122.29 | 131.61 |

**Table 4.** Total energy consumed at round 5000 in WSN #3 with 50 CHs.

| Number of Sensors = 500 | BS center | BS corner | BS outfield |
|---|---|---|---|
| Direct Transmission (DT) | 250.00 | 250.00 | 250.00 |
| LEACH | 250.00 | 250.00 | 250.00 |
| LEACH -C | 235.62 | 245.91 | 250.00 |
| PSO - C | 224.83 | 230.74 | 248.16 |
| WOA - C | 202.35 | 210.24 | 217.45 |

**Network lifetime performance:** Next, the algorithm was run for comparison of a lifetime under different conditions. Network lifetime is taken as the last round until which at least 1 node is alive. The total number of sensor nodes varied between 100-500 and the number of CH varied from 10 to 50. It can be seen from Fig. 8 that WOA-C gives much better lifetime compared to LEACH and is on par with PSO-C.

The reason for PSO-C and WOA–C giving better performance is that these algorithms use better CH selection process which involves considering the residual energy of a node before selecting it as a CH.

The DT has a network lifetime of 732, LEACH has a lifetime of 2747 rounds while LEACH-C lasts for around 4755 rounds. PSO-C performs better due to the restricted selection of CHs and lasts for about 6639 rounds and lastly, WOA-C has a network lifetime of 7268 clearly outperforming other algorithms.

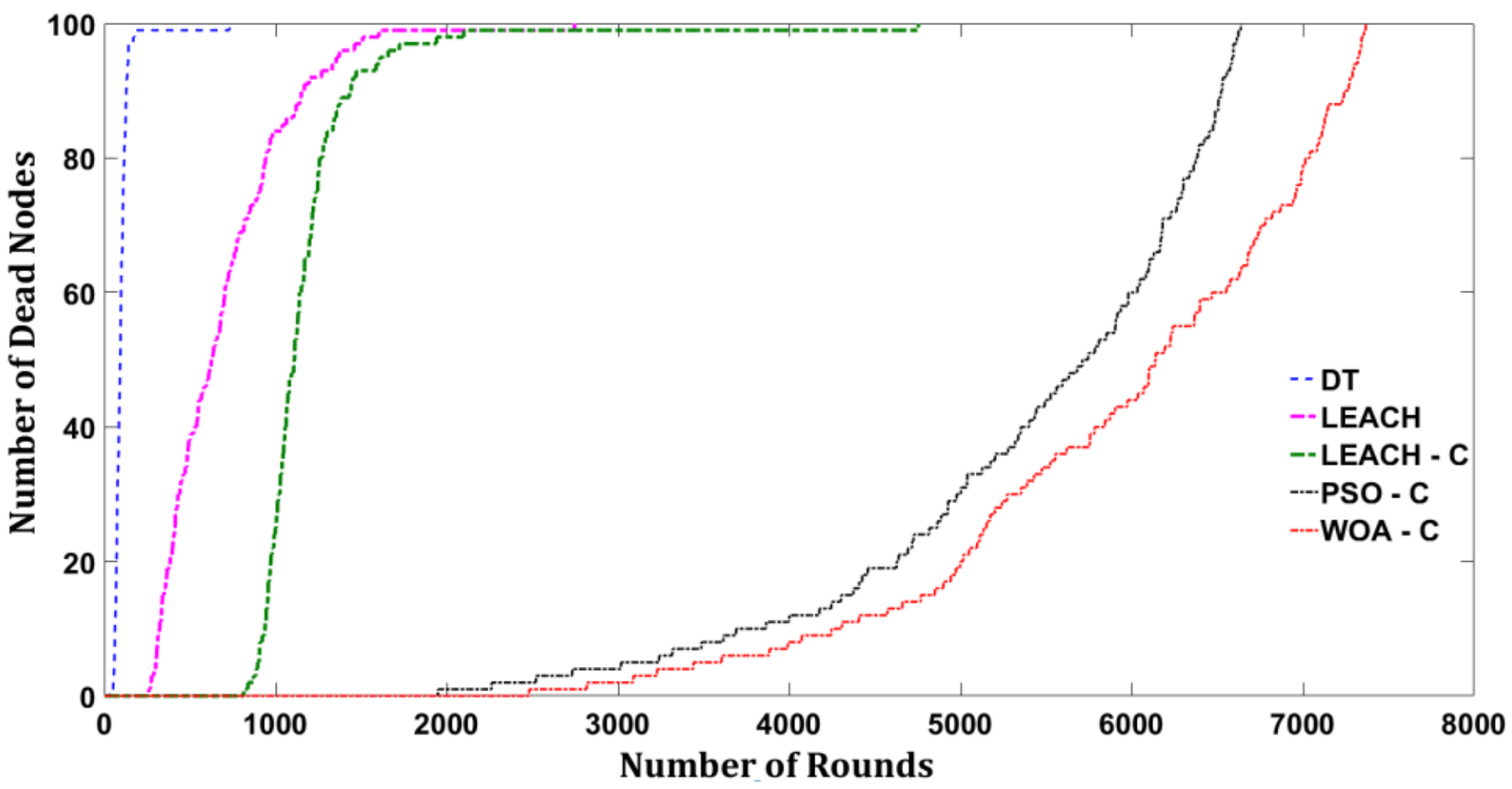

**Fig. 8.** Network lifetime performance in WSN#1 with 100 nodes and 10CHs

It can also be noted from Fig. 9, 10, 11 that the network lifetime is highest with BS at a central position and least with outfield position. Because the CHs need to transmit over greater distances when BS is outside.

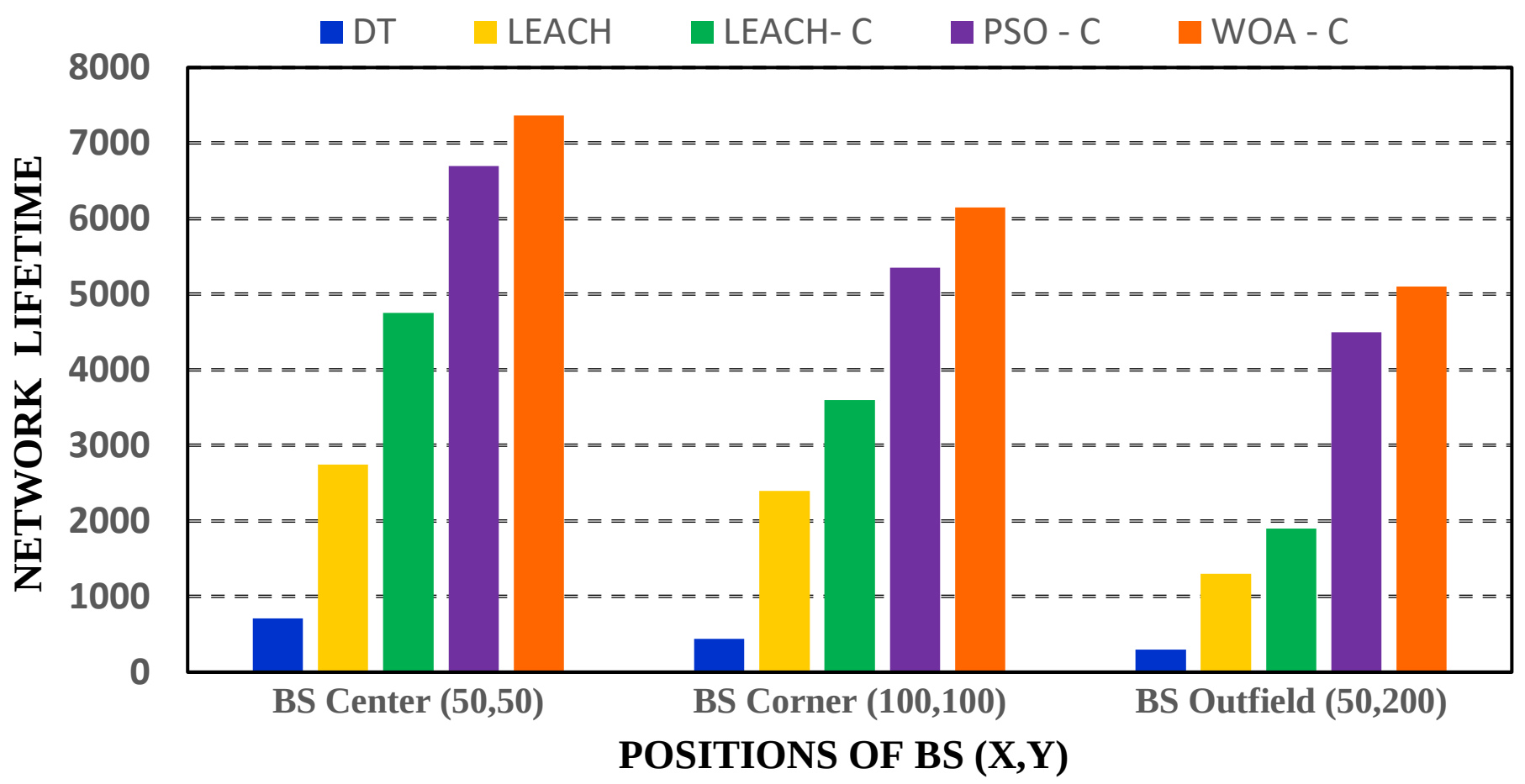

**Fig. 9.** Comparisons of network lifetime with varying BS positions in WSN#1 with 10 CHs.

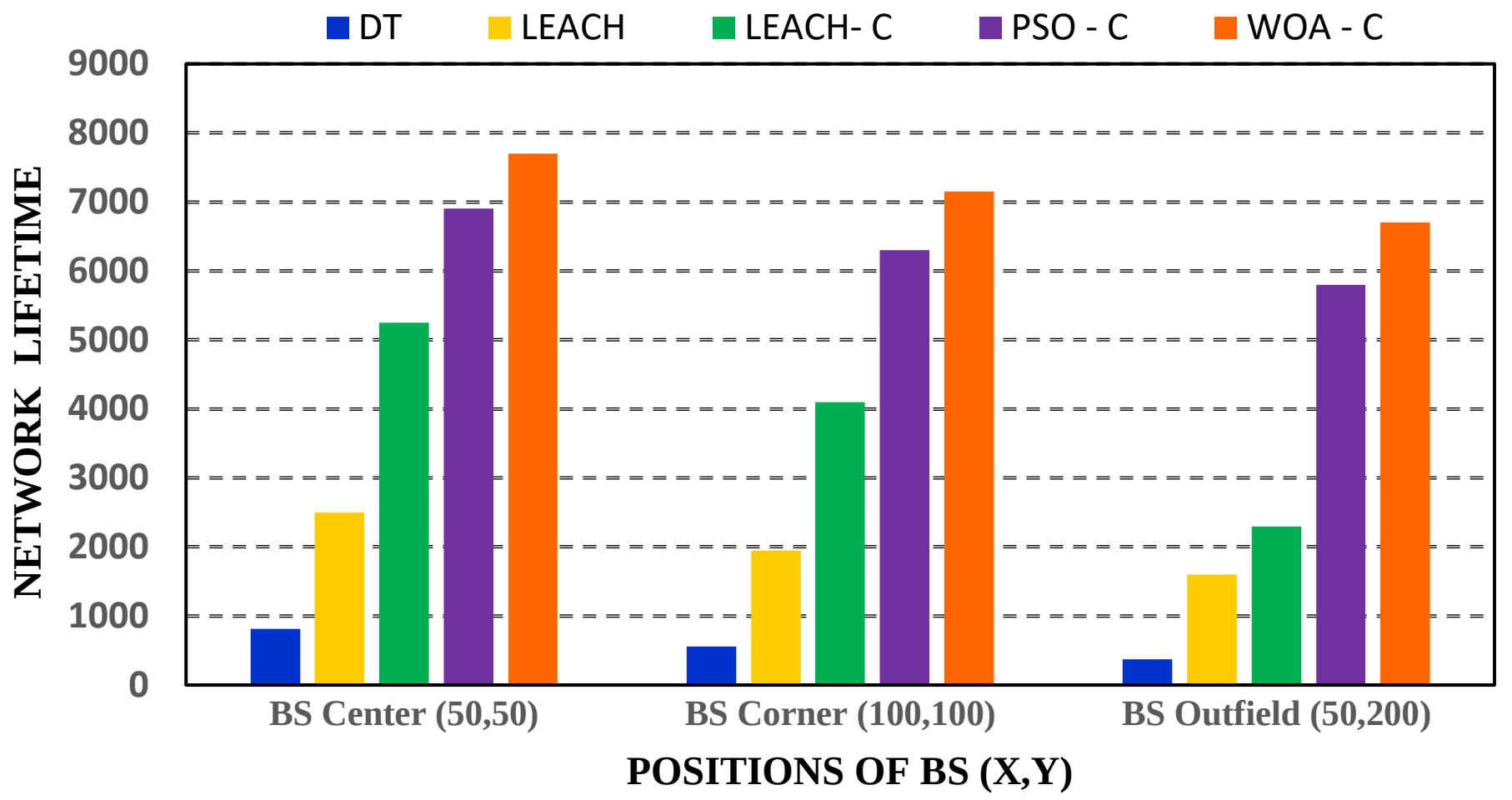

**Fig. 10.** Comparisons of network lifetime with varying BS positions in WSN#2 with 30 CHs.

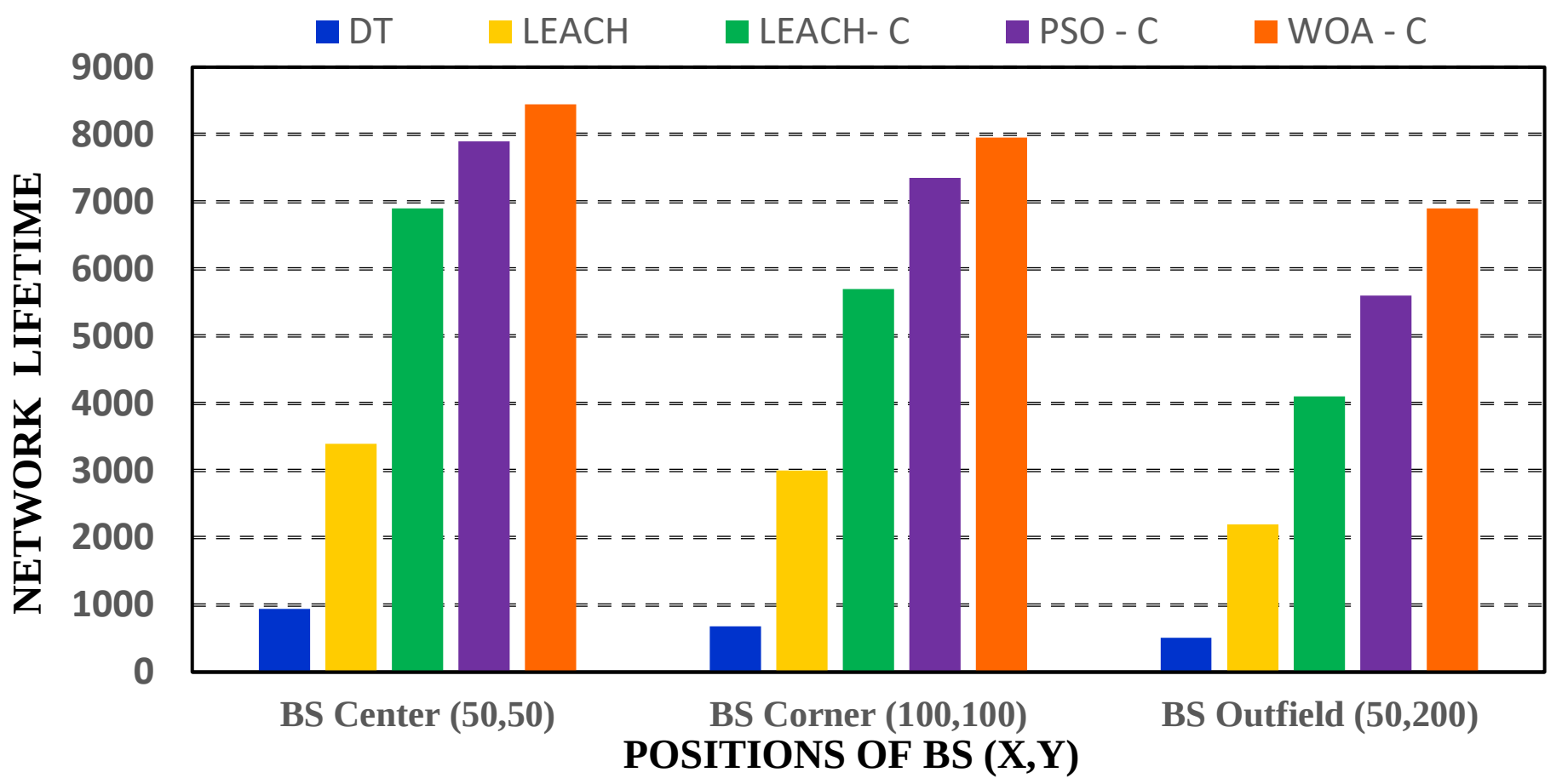

**Fig. 11.** Comparisons of network lifetime with varying BS positions in WSN#3 with 50 CHs.

As shown in the above figures, WOA-C consistently performs better even under varying conditions such as changes in a number of cluster heads, network size, and different BS positions.

**Throughput comparison**. Finally, the simulations are run for comparing throughput across different algorithms. Throughput is defined as the total number of packets received by the BS at a particular instant of time. Throughput is calculated for WSN#1 with 100 nodes and 10 CHs whose result is shown in Fig. 12. Since the proposed algorithm uses lesser energy, it has more network lifetime. This implies that the pro-posed algorithm has a higher number of live nodes in the network at any given point of time when compared to other algorithms. Hence, the throughput of the proposed algorithm is higher than the remaining algorithms although the performance of PSO-C comes close. The throughput of the network with more nodes will be higher due to more number of sensors. Moreover, the proposed algorithm is stable even when BS positions are changed. The throughput of the network corresponds to the total number of live nodes in the network and hence the network in which BS is centrally positioned usually gives a better throughput. The main point of differentiation of PSO and WOA based clustering is that these algorithms consider residual energy and number of neighboring nodes while selecting CHs. The overall performance of the algorithm is given by Table 5 and the throughput is taken at round number 2000.

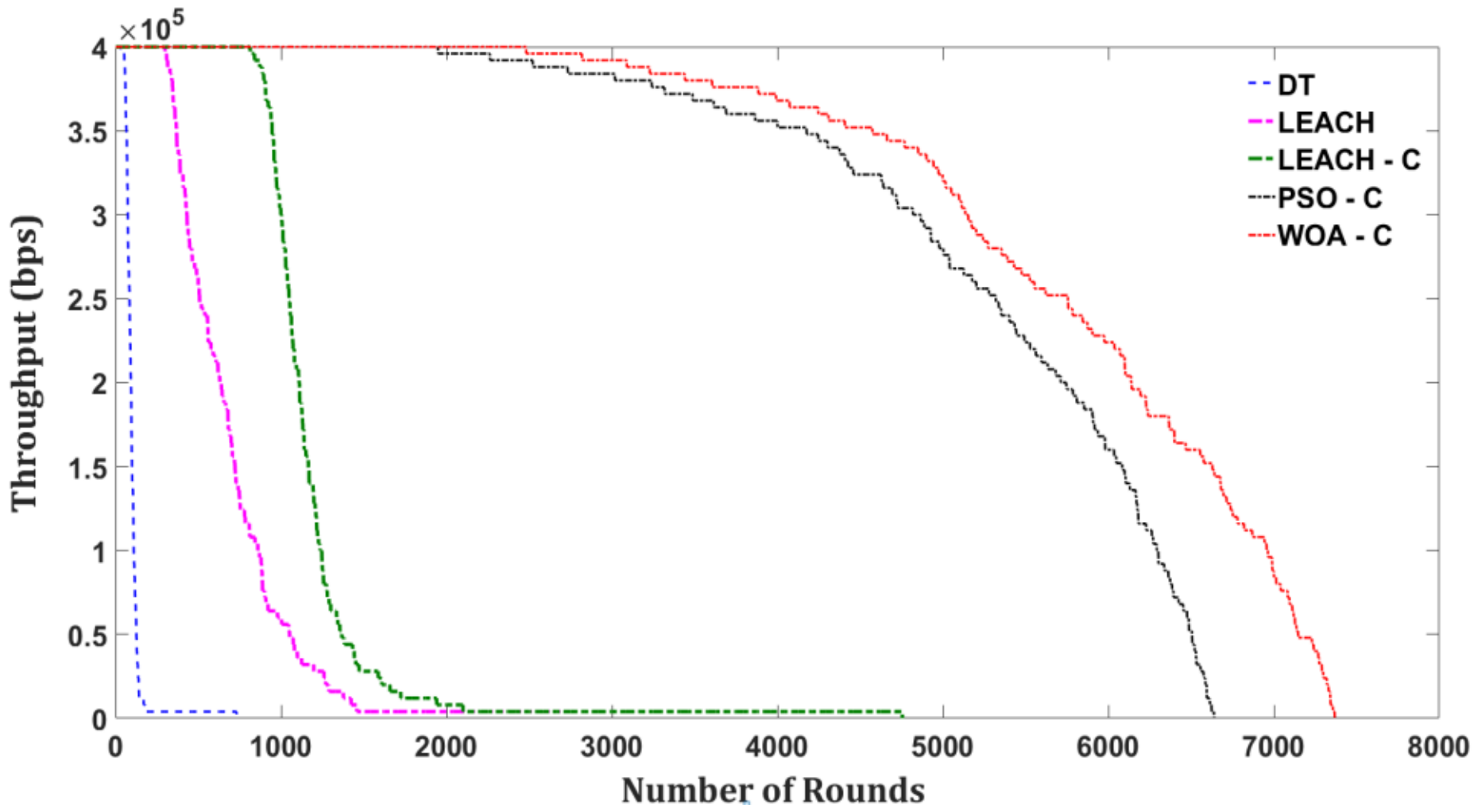

**Fig. 12.** Throughput comparison in WSN#1 with 100 nodes and 10CHs

The last node death, first node death, and throughput at round 2000 are given below for various algorithms simulated on WSN#1.

**Table 5.** Overall Performance of WSN#1 with 100 nodes and 10CHs.

| Number of Sensors = 100 | FND | LND | Throughput at round 2000 (bits /round) |
|---|---|---|---|
| Direct Transmission (DT) | 31 | 732 | 0 |
| LEACH | 260 | 2747 | 4000 |
| LEACH -C | 814 | 4755 | 8000 |
| PSO - C | 1949 | 6639 | 396000 |
| WOA - C | 2482 | 7268 | 400000 |

Conclusively, the proposed algorithm was tested under many different conditions and in each of this scenario the WOA-C algorithm outperformed all the other standard routing protocols. This is due to the implementation of whale optimization on an efficient fitness function in the proposed algorithm designed to consider many factors such as residual energy and, the number of neighboring nodes, before considering a node for cluster head. The total energy consumption and network lifetime of the network are highest when the BS is centrally positioned and when the BS is located outside the sensing area then the performance is lower.

## 6 Conclusions

In this paper, an energy efficient routing algorithm based on whale optimization algorithm (WOA) called WOA-C is presented. It involves selection of energy aware cluster heads based on fitness function which considers the residual energy of the node, and the sum of energy of adjacent nodes decreasing the overall energy consumption of the sensor network. In addition, the performance of WOA-C is evaluated against other standard contemporary routing protocols such as LEACH, LEACH-C, and PSO-C. The performance of various algorithms is evaluated for energy efficiency, network lifetime and throughput. Consequently, extensive simulations show that the proposed algorithm outperforms other routing algorithms in terms of residual energy, network lifetime, throughput and longer stability period. In conclusion, this study presents a robust energy efficient routing algorithm based on Whale optimization, which has the ability to select cluster heads for maximum energy utilization in WSN. It has been shown that the proposed method outperforms other contemporary routing protocols.

Future works will aim to develop a routing algorithm more efficient by considering more factors in the fitness function like energy balancing. An algorithm that works for heterogeneous WSN can also be developed. The optimization can also be improved by using hybrid optimization techniques to increase the search efficiency in order to converge at optimal solutions quickly.